\pdfoutput=1
\documentclass[10pt,twocolumn,letterpaper]{article}
\usepackage[pagenumbers]{cvpr}
\usepackage{booktabs}
\usepackage{graphicx}
\usepackage{amsmath}
\usepackage{xspace}
\newcommand{\HL}{\mbox{\textsc{Baton}}\xspace}%
\usepackage{colortbl}
\usepackage{amssymb}
\definecolor{oursrow}{HTML}{E9F2FA}
\newcommand{\ci}[1]{{\scriptsize\textcolor{black!55}{[#1]}}}
\newcommand{\grp}[1]{\textcolor{black!60}{\emph{#1}}}
\newlength{\vaw}\newlength{\vbw}
\newcommand{\rb}[1]{\makebox[\vaw][r]{#1}}
\newcommand{\an}[1]{\hspace{-5.5pt}\makebox[\vbw][l]{#1}}

\newcommand{\phead}[2][8pt plus 4pt minus 1pt]{\par\vspace{#1}\noindent\textbf{#2}\hspace{1em}\ignorespaces}
\definecolor{cvprblue}{rgb}{0.21,0.49,0.74}
\usepackage[pagebackref,breaklinks,colorlinks,allcolors=cvprblue]{hyperref}

\title{Perceive Once, Escalate on Demand:\\Handing Off Visual KV Memory from Small to Large VLMs}
\makeatletter\patchcmd{\@maketitle}{\vspace*{24pt}}{\vspace*{17pt}}{}{\errmessage{title patch failed}}\makeatother\newcommand{\authordrop}{10.5pt}
\author{\fontsize{10}{12.5}\selectfont \textbf{Qiang Zhou}\textsuperscript{1,*} \quad \textbf{Jianing Deng}\textsuperscript{1} \quad \textbf{Siyan Dong}\textsuperscript{2} \quad \textbf{Jingxian Chen}\textsuperscript{1} \quad \textbf{Han-Yi Lin}\textsuperscript{1} \quad \textbf{Jingtong Hu}\textsuperscript{1,\dag}\\[3.5pt]
{\fontsize{9.5}{11}\selectfont \textsuperscript{1}University of Pittsburgh \qquad \textsuperscript{2}University of Hong Kong}\\[-\authordrop]
}
\makeatletter
\def\ps@firstpage{\let\@oddhead\@empty\let\@evenhead\@empty
  \def\@oddfoot{\setbox0\hbox{\footnotesize \textsuperscript{\dag}Corresponding author.}%
    \rlap{\raisebox{18pt}{\rule{\wd0}{0.4pt}}}\rlap{\raisebox{9.5pt}{\footnotesize \textsuperscript{*}First author.}}\rlap{\box0}\hfil\thepage\hfil}\let\@evenfoot\@oddfoot}
\makeatother

\begin{document}
\maketitle
\thispagestyle{firstpage}
\begin{abstract}
An embodied agent cannot afford a large vision-language model (VLM) at every step of a long video, so a cascade runs a small VLM continuously and escalates only uncertain decisions to a large one.
Each escalation, however, forces the large model to re-encode a history the small model has already perceived, and this dominates its cost.
We present \HL, which hands the large model the small model's visual memory, its key-value (KV) cache over the history.
\HL rests on one finding: the large model needs a \emph{usable} memory, not a faithful copy of its own.
A map trained to reproduce the large model's cache matches it most closely, yet brings no gain over the small model alone.
\HL instead distills a per-layer linear map, without labels, from the large model's own answers, with the large model encoding only the current frame.
Interventions on the map trace the gain to one direction, carried by the middle third of the large model's layers.
On RoboMME, \HL beats the small model alone, a pruned or shortened re-watch, and a text memory; as the escalation channel of a cascade it matches re-watching to within about one point at 28--47\% less compute, and 8$\times$ less per escalation at 64 frames.
Its gain grows with the large model, from 8B to 32B, tracks the large model's headroom across eight model pairs from four families, and transfers zero-shot to OpenEQA.
\end{abstract}
\section{Introduction}
\label{sec:intro}

\begin{figure}[t]
  \centering
  \includegraphics[width=\linewidth]{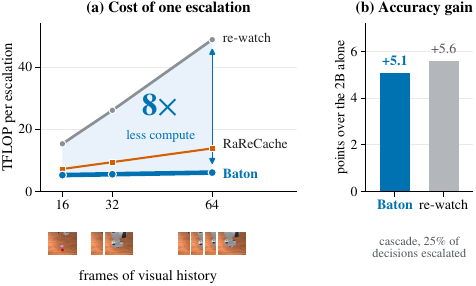}
  \caption{\textbf{Escalation without re-watching.} (a) Compute of the large VLM (Qwen3-VL 8B) for one escalation vs.\ history length: re-watching grows linearly, RaReCache~\cite{sarkar2026rarecache} more slowly, and \HL stays nearly flat, 8$\times$ lower at 64 frames. (b) With 25\% of decisions escalated, \HL keeps nearly all the accuracy that re-watching adds (\cref{sec:exp_cascade}).}
  \label{fig:cost_scale}
\end{figure}

\begin{figure*}[t]
  \centering
  \includegraphics[width=\linewidth]{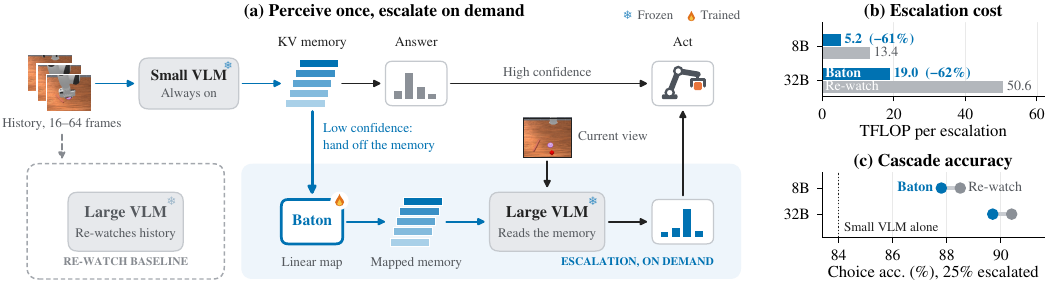}
  \caption{\textbf{Perceive once, escalate on demand.} (a) A small VLM runs continuously over the agent's visual history and keeps its KV cache. When it is unsure, the decision is escalated: instead of re-encoding the whole history, the large VLM reads the small model's memory through \HL, a per-layer linear map, and encodes only the current view itself. (b) Reader compute per escalation on 16-frame histories. (c) At a 25\% escalation budget the memory cascade is within 0.7 points of re-watching for both readers.}
  \label{fig:overview}
  \vspace{-2.6pt}%
\end{figure*}

An embodied agent that acts over minutes must remember what it saw: where an object was hidden, which button was pressed, what changed since the last step.
Running a large vision-language model (VLM) over the whole visual history at every decision is too expensive, so a cascade runs a small VLM continuously and escalates only the decisions it is unsure of to a large one~\cite{chen2023frugalgpt,ong2025routellm,gupta2024cascades}, much as a robot asks an expert for help when it is uncertain~\cite{ren2023knowno,singh2022ask4help}.

Escalation has a hidden cost.
The large model has never seen the history, so it must re-encode every frame before it can answer, and this re-encoding, not the answer, dominates its compute (\cref{fig:cost_scale}): at 64 frames it takes nine tenths of an escalation.
Yet the small model has already perceived the same history: its key-value (KV) cache over the video is a visual memory that exists for free at the moment of escalation.
A transformer reads its past only through this cache: later tokens attend to its keys and values alone.
Describing that memory in words loses most of it; the cache keeps what the small model saw, in a form another model can attend to.
This raises a critical question (\cref{fig:overview}): \textbf{can the large model read that memory instead of watching again?}

The natural way to hand over a cache is to translate it: fit a map from the small model's KV to the large model's own KV over the same history, as has been done between text language models~\cite{heo2026crossmodel,sarkar2026rarecache}.
On video this fails in an instructive way.
The fitted maps reproduce the large model's cache more closely than any other map we train, and a large model reading them answers no better than the small model alone.
What it reads from them is almost what it reads from its own cache, and it still gains nothing.
Matching a representation and matching its function are different things~\cite{csiszarik2021similarity}, and here the difference is the whole effect.

\textbf{We find that the large model needs a \emph{usable} memory, not a faithful copy of its own.}
\HL therefore does not regress the cache.
It distills a per-layer linear map, without labels, from the large model's own answers, through the path in which the map is deployed: the history comes from the small model's memory and the large model encodes only the current frame.
The resulting memory is farther from the large model's native cache and carries the whole gain.
By what the large model reads from it, it is eleven times farther than the fitted maps.

Because the map is linear, we can test this directly.
A random step with exactly the norm of the gap between the fitted and the distilled map costs nothing; the same step along the gap costs the whole gain.
The gain sits in the middle third of the large model's layers, which alone are necessary and sufficient, in two model families.
The correction is mostly in the keys, in where the large model looks, and sits where the distilled map departs least from the large model's own cache.
Handing off a memory is thus not a reconstruction problem but a search, among many nearby memories, for one the reader can use, and only the reader's own answers say which (\cref{sec:exp_mech}).

On RoboMME~\cite{robomme}, a large model reading the small model's memory beats the small model alone, a closed-form KV regression, a pruned or shortened re-watch~\cite{chen2024fastv,alvar2025divprune}, and a text memory.
As the escalation channel of a cascade, \HL matches re-watching to within about one point at 28--47\% less compute, and its cost stays nearly flat as the history grows.
At 64 frames an escalation needs 8$\times$ less compute and runs 2.6$\times$ faster on one RTX 4090.
The gain grows with the large model, from 8B to 32B, tracks the large model's headroom across eight model pairs from four families, and transfers zero-shot to OpenEQA~\cite{majumdar2024openeqa}.

In summary, we (1) pose the handoff of visual KV memory as the escalation channel of a small-to-large VLM cascade (\cref{sec:setup}), (2) introduce \HL, a label-free linear handoff distilled from the large model's answers (\cref{sec:method}), and (3) show by intervention why a usable memory, not a faithful one, is what the large model needs (\cref{sec:exp_mech}).
\section{Related Work}
\label{sec:related}
\newcommand{\rwhead}[2][0pt]{\par\vspace{#1}\noindent\textbf{#2}\hspace{1em}\ignorespaces}

\rwhead{Cascades, routing and deferral.}
Running a cheap model first and a costly one only when needed is an old idea in vision~\cite{viola2001rapid,teerapittayanon2016branchynet}.
For language models, FrugalGPT~\cite{chen2023frugalgpt} and its successors learn when to escalate from self-verification~\cite{aggarwal2024automix}, query difficulty~\cite{ding2024hybridllm,ong2025routellm} or token uncertainty~\cite{gupta2024cascades}; SGL~\cite{zhao2025sgl} does so for VLMs, and embodied agents ask a human~\cite{ren2023knowno} or an expert policy~\cite{singh2022ask4help}.
All of them decide \emph{when} to escalate, and the large model then starts from the raw input, which for a long video history is the dominant cost.
We study \emph{what is handed over}; any of these routers can replace our confidence threshold.

\rwhead[4.5pt]{Cross-model KV transfer.}
The closest work maps KV caches between text LLMs of different sizes so that the receiver skips prefill: Heo~\etal~\cite{heo2026crossmodel} fit a closed-form per-head ridge map and motivate it with cascades without evaluating one; RaReCache~\cite{sarkar2026rarecache} also recomputes about 30\% of the positions; Cache-to-Cache~\cite{fu2025c2c}, KVComm~\cite{kvcomm2025} and DroidSpeak~\cite{liu2024droidspeak} share or fuse caches.
All are text-only and fit the map to the reader's KV; we hand over a \emph{video history}, distill the reader's answers through the deployed path instead of regressing KV, and evaluate the cascade end to end.
That fitting activations and training on the task loss give different layers is known from model stitching~\cite{bansal2021revisiting,csiszarik2021similarity}; here the gap decides the outcome, and we localize it.
Both approaches are baselines (\cref{sec:exp_handoff} and \cref{sec:exp_alternatives}).

\rwhead[4.5pt]{Cheaper re-watching and memory within one model.}
Token reduction merges~\cite{bolya2023tome} or drops visual tokens~\cite{chen2024fastv,zhang2025sparsevlm,yang2025visionzip}, but a pruned re-watch still runs the large model's vision tower over every history frame.
Streaming video models keep a KV memory of past frames~\cite{chen2024videollmonline,di2025rekv,xu2026streamingvlm} and embodied agents a text memory~\cite{anwar2024remembr}, each within one model.
Our reader never encodes the history: it reads a memory that a smaller model has built, and we compare against pruned re-watching and text memory (\cref{sec:exp_alternatives}).
\section{Problem Setup}
\label{sec:setup}

\phead[0pt]{Writer and reader.}
A small \emph{writer} $W$ and a large \emph{reader} $R$ come from the same model family (here Qwen3-VL~\cite{qwen3vl} 2B and 8B) and share tokenizer, image preprocessing and positions.
At a decision step the input is a visual history $H$, the current observation and a question, rendered as one token sequence.
$W$ prefills the history prefix and keeps its KV cache; this cache is the \emph{memory}.
A handoff map $f$ turns $W$'s cache into one $R$ can read, and $R$ answers from it without seeing the history frames.

\phead[8pt plus 4pt minus 1pt]{When a handoff pays off.}
\emph{Accuracy:} $R$ reading $f(\text{memory})$ must beat $W$ answering from its own memory ($W$-self), otherwise escalation buys nothing.
\emph{Cost:} the handoff must cost less than $R$ re-encoding the history itself ($R$-native), otherwise $R$ might as well re-watch.
We also report $R$ with no history, the floor showing that the memory is used, and the share of $R$'s advantage the handoff retains,
\begin{equation}
  \text{retention} = \frac{\text{acc}(\text{handoff}) - \text{acc}(W\text{-self})}{\text{acc}(R\text{-native}) - \text{acc}(W\text{-self})}.
\end{equation}

\phead[8pt plus 4pt minus 1pt]{Memory cascade.}
$W$ answers every decision with a confidence (choice: the margin between its two most probable options; point: the mean token log-probability of its answer), and a decision is escalated when the confidence falls below a threshold set on validation data for an escalation budget $b$ (10--40\% of decisions).
An escalated decision is answered by $R$ through one of two channels: \emph{re-watch} ($R$ prefills the whole history) or \emph{memory} ($R$ reads $f(W\text{'s cache})$ and natively encodes only the most recent frames).
Both cascades share writer answers, router and budget, so their difference isolates the channel.
Cost is TFLOP per decision: the writer's prefill is always paid, and escalation adds $R$'s full prefill or the map and $R$'s suffix.
\section{A Faithful Memory Is Not a Usable One}
\label{sec:diag}

\begin{figure*}[t]
  \centering
  \includegraphics[width=\linewidth]{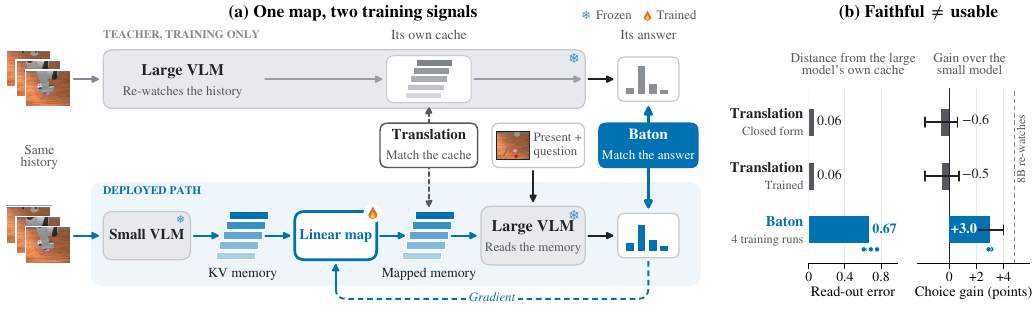}
  \caption{\textbf{Two ways to train the same handoff map, and what they give} (2B$\to$8B, RoboMME choice). (a) The small model's KV memory is mapped into a memory the large model reads while it encodes the present itself. \emph{Translation} fits it to the large model's own cache; \HL distills the large model's answer through the deployed path. (b) The translation maps are eleven times closer to the large model's cache than \HL and gain nothing; \HL carries the gain in each of four runs (dots; bars: mean, 95\% interval).}
  \label{fig:diag}
\end{figure*}

The natural way to build the handoff map $f$ is to \emph{translate}: fit $W$'s cache to the cache $R$ would have computed over the same history, as is done between text LLMs~\cite{heo2026crossmodel,sarkar2026rarecache}.
We give it every advantage we can, and it does not pay off.

\phead[8pt plus 4pt minus 1pt]{A per-layer linear map.}
Writer and reader differ in depth (28 vs.\ 36 layers) but share 8 KV heads of dimension 128.
For reader layer $l$ we concatenate the key heads (or value heads) of the two writer layers $a(l)$ and $b(l)$ nearest in relative depth with a bias, and map the 2049-d vector linearly, with weights $A^{X}_{l}$, to the reader's 1024-d keys (or values):
\begin{equation}
  \hat{X}^{R}_{l} = \big[X^{W}_{a(l)} \,\|\, X^{W}_{b(l)} \,\|\, 1\big]\, A^{X}_{l}, \qquad X \in \{K, V\}.
  \label{eq:map}
\end{equation}
Keys and values have separate weights, every layer is independent, and the rotary position encoding of the keys is removed before the map and re-applied after it.
Every map in this paper has this form; the maps differ only in how the weights are obtained.

\phead[8pt plus 4pt minus 1pt]{The reader looks at the present itself.}
Only the history prefix is mapped.
The current observation and the question are prefilled natively by the reader on top of the mapped history; they sit at the end of the sequence, so this step is causally exact.
We call this the \emph{hybrid path}, and it is the deployed form throughout: the small model remembers, and the large model looks at what is in front of it.

\phead[8pt plus 4pt minus 1pt]{Fitting the reader's cache.}
Translation regresses $R$'s native KV on $W$'s KV over calibration videos, with no questions or answers; with $Z_l$ the bracket of \cref{eq:map} stacked over the calibration positions and $X^{R}_{l}$ the reader's native keys or values there, each layer has the closed-form ridge solution $A^{X}_{l} = (Z_l^{\top} Z_l + \alpha_l D)^{-1} Z_l^{\top} X^{R}_{l}$~\cite{heo2026crossmodel}, where $D$ is the identity without its bias entry and the penalty $\alpha_l$ is set on held-out videos.
To rule out a problem of optimization, we also train the map by gradient descent on the KV error alone, through the hybrid path, with the initialization, data and number of steps of our method (\cref{sec:method}).

\phead[8pt plus 4pt minus 1pt]{The most faithful maps gain nothing.}
Both maps do what they were fit to do.
On 200 test decisions we hold the reader's native queries fixed, replace only the history rows of its cache by the mapped ones, and measure the relative error of each layer's attention output, the \emph{read-out error}.
From the fitted maps the reader reads almost what it reads from its own cache (read-out error 0.06), with almost the same attention pattern, and no other map we train comes close to this fidelity (\cref{fig:diag}b).
Yet the 8B reading them answers no better than the 2B answering alone: 83.4 choice against 84.0 for both maps ($-$0.6 [$-$1.8, 0.6] and $-$0.5 [$-$1.8, 0.7]), and 79.2 and 79.0 point against 80.5.
The mapped memory is read, since the same 8B with no history scores 81.3, but it gives the large model less than the small model takes from its own memory, so by the accuracy condition of \cref{sec:setup} the escalation buys nothing.
Nor is the failure specific to this pair: the closed-form map does not beat its writer with a 32B reader (+1.1 [$-$0.4, 2.6]) or in a second model family (\cref{sec:exp_scale}).

\phead[8pt plus 4pt minus 1pt]{Hypothesis.}
A cache close to the reader's own is still not the reader's own, and what the residual error does depends on how the reader uses the memory, which a regression target never sees.
We therefore hypothesize that \emph{the reader does not need its own memory back; it needs a memory it can use}: among maps that fit the reader's KV about equally well, some should carry the reader's advantage and others none, and only the reader's answers can tell them apart.
\Cref{sec:method} trains the map on this signal, and \cref{sec:exp_mech} tests the hypothesis by intervention.
\section{\HL: Distilling a Usable Memory}
\label{sec:method}

\HL keeps the map of \cref{eq:map} and the hybrid path, and changes only the training signal: it is distilled from the reader's answers, through the deployed path (\cref{fig:diag}a).

\phead[8pt plus 4pt minus 1pt]{Teacher and student.}
The teacher is the reader on its own native cache of the history; the student is the same frozen reader on the mapped history plus its natively prefilled suffix, exactly the test-time path.
For each decision the teacher greedily answers a probe, and the map minimizes
\begin{align}
  &\mathcal{L} = \mathrm{KL}\big(p_{\text{teacher}} \,\|\, p_{\text{student}}\big) + \lambda\, \mathcal{L}_{\text{KV}}, \qquad \lambda = 0.1, \label{eq:loss}\\
  &\mathrm{KL}\big(p_{\text{teacher}} \,\|\, p_{\text{student}}\big) = \tfrac{1}{2}\, \delta^{\top} F\, \delta + O(\|\delta\|^{3}), \label{eq:fisher}
\end{align}
a full-vocabulary KL on the teacher's response plus a normalized mean-squared KV error on the history.
In \cref{eq:fisher}, $\delta$ is the residual of the mapped history against the reader's own cache and $F$ the Fisher information of the reader's answer with respect to its memory: regression measures only the size of $\delta$, distillation what $\delta$ does to the answer.
Gradients reach the map through the reader's attention; writer and reader stay frozen.
The map starts from the closed-form solution, and we call this handoff \HL.

\phead[8pt plus 4pt minus 1pt]{No labels.}
No annotation is used: the probes are the decision questions themselves on RoboMME, with the teacher's own answer as the target, and generic scene-description prompts on VSI-Bench~\cite{yang2025thinking} and OpenEQA; training uses 1500 unlabeled decisions for two epochs.

\phead[8pt plus 4pt minus 1pt]{The training path must be the deployed path.}
A map distilled with every position mapped and then moved into the hybrid path gets worse, not better (\cref{sec:exp_handoff}): it was trained for a reader whose view of the present is mapped as well.
The student is therefore the hybrid path itself.

\phead[8pt plus 4pt minus 1pt]{Hard-decision training.}
A cascade escalates only the decisions the writer is unsure about.
For the cascades we therefore continue training \HL, still label-free, on the 25\% of training decisions with the lowest writer confidence, with a replay of ordinary decisions.
\section{Experiments}
\label{sec:experiments}

\subsection{Setup}
\label{sec:exp_setup}

\paragraph{Models and benchmarks.}
The main pair is Qwen3-VL 2B$\to$8B.
\textbf{RoboMME}~\cite{robomme} is a benchmark of 16 robot-manipulation tasks that test temporal, spatial, object and procedural memory. We test on episodes 0--29 of all tasks (1647 choice and 1306 point decisions) and train and validate on the rest; histories are 16 key frames or up to 64 frames.
\emph{Choice} asks which action to take next; \emph{point} asks where to apply a given action (hit if within 16\,px).
Zero-shot, the 8B does not use RoboMME history at all, so writer and reader are each LoRA-fine-tuned~\cite{hu2022lora}; the handoff is between the fine-tuned models, and the map never sees a label.
\textbf{OpenEQA} EM-EQA~\cite{majumdar2024openeqa} asks open-ended questions about an indoor scene seen in a video tour. The base models are used zero-shot; every design choice is made on the HM3D~\cite{ramakrishnan2021hm3d} episodes (557 questions), and the ScanNet~\cite{dai2017scannet} episodes (1079 questions) are held out.
Answers are scored with the official LLM-Match prompts by a local Qwen3-32B judge and re-scored with GPT-4o.

\paragraph{Compute and statistics.}
TFLOP are counted analytically.
Every comparison is a paired difference with a 95\% episode-cluster bootstrap interval.
Each experiment's hypotheses, gates, splits and multiplicity correction were written to a protocol file before it ran.

\subsection{The large model can read the memory}
\label{sec:exp_handoff}

\begin{table}[t]
  \caption{\textbf{Handoff on RoboMME (2B$\to$8B, 16-frame history).} Translation fits the map to the reader's cache (\cref{sec:diag}), distillation to its answers; in the hybrid path the reader encodes the present itself. Only \HL beats the 2B alone on choice (+2.9 [1.6, 4.3]); point +1.9 [$-$0.2, 4.1] (95\% intervals, Bonferroni-corrected).}
  \label{tab:handoff}
  \centering\small
  \setlength{\tabcolsep}{5pt}
  \begin{tabular}{@{}lcccc@{}}
    \toprule
    & \multicolumn{2}{c}{\grp{hybrid path in}} & & \\
    \cmidrule(lr){2-3}
    RoboMME test & \grp{training} & \grp{test} & choice & point \\
    \midrule
    2B alone ($W$-self) & & & 84.0 & 80.5 \\
    8B, current frame only & & & 81.3 & 74.4 \\
    8B re-watches ($R$-native) & & & 88.8 & 83.1 \\
    \midrule
    \multicolumn{5}{@{}l}{\grp{8B reads the 2B's memory through a map}} \\
    Translation & & & 75.3 & 75.7 \\
    Translation~\cite{heo2026crossmodel} & & \checkmark & 83.4 & 79.2 \\
    Distillation & & & 84.5 & 79.9 \\
    Distillation & & \checkmark & 83.1 & 78.7 \\
    \rowcolor{oursrow} \textbf{\HL} (distillation) & \checkmark & \checkmark & \textbf{86.9} & \textbf{82.4} \\
    \bottomrule
  \end{tabular}
\end{table}

On RoboMME, the 8B reading the 2B's memory beats the 2B answering alone on choice, +2.9 [1.6, 4.3] (\cref{tab:handoff}); on point it is +1.9 [$-$0.2, 4.1].
Over four training runs choice accuracy is 86.8--87.1 and the gain is significant in every run (mean +3.0 [1.9, 4.0]); point accuracy varies more (80.6--82.4, mean gain +1.0 [$-$0.7, 2.7]).
It recovers 61\% of the 8B's choice and 74\% of its point advantage over the 2B, and it reads the memory, not just the current frame: +5.6 choice and +8.0 point above the 8B with no history.

The reader must encode the current frame itself (\cref{tab:handoff}): mapping it too costs the closed-form map 8 choice points.
Translation does not beat the writer even in the hybrid path, and \HL beats it by +3.5 [2.4, 4.7] on choice and +3.2 [1.8, 4.7] on point.
The map must be trained through the path it is deployed in: a map distilled with the present mapped as well gets worse once the reader encodes the present itself ($-$1.3 [$-$2.4, $-$0.3]), while distilling through the hybrid path gives \HL.
On the seven memory-dependent tasks \HL reaches 95.1 choice (writer 91.4, reader 97.0), +7.9 [5.6, 10.2] above translation.

\subsection{Re-watching less costs more and loses}
\label{sec:exp_alternatives}

\begin{table}[t]
  \caption{\textbf{Alternatives to the memory handoff} (2B$\to$8B, RoboMME test, 16-frame histories; TFLOP per escalation). Every other way of seeing the history costs more and loses on choice or point; text memory loses on both, even with a trained reader.}
  \label{tab:alternatives}
  \centering\small
  \setlength{\tabcolsep}{5pt}
  \settowidth{\vaw}{10.55}
  \begin{tabular}{@{}lccc@{}}
    \toprule
    How the 8B gets the history & choice & point & TFLOP$\downarrow$ \\
    \midrule
    \multicolumn{4}{@{}l}{\grp{Watches less}} \\
    4 frames & 82.5 & 80.8 & \rb{7.04} \\
    16 frames, half resolution & 86.7 & 78.3 & \rb{7.60} \\
    8 frames & 86.3 & 82.0 & \rb{9.15} \\
    \multicolumn{4}{@{}l}{\grp{Re-watches with pruned history tokens}} \\
    10\% kept (DivPrune~\cite{alvar2025divprune}) & 83.8 & 77.6 & \rb{7.93} \\
    50\% kept (uniform) & 87.6 & 82.0 & \rb{10.37} \\
    50\% kept (FastV~\cite{chen2024fastv}) & 88.0 & 80.0 & \rb{10.55} \\
    \multicolumn{4}{@{}l}{\grp{Reads the 2B's caption of the history (text memory)}} \\
    reader zero-shot & 75.8 & 72.1 & \rb{7.34} \\
    reader fine-tuned on captions & 83.2 & 76.6 & \rb{7.34} \\
    \midrule
    \rowcolor{oursrow} \textbf{\HL} (reads the 2B's memory) & \textbf{86.9} & \textbf{82.4} & \rb{\textbf{5.21}} \\
    \bottomrule
  \end{tabular}
\end{table}

Every lighter way for the 8B to look at the history costs more than \HL and loses on choice or point (\cref{tab:alternatives}).
Token pruning cannot remove the dominant cost: a pruned re-watch must still run the 8B's vision tower over every history frame, so even keeping \emph{none} of the history tokens costs 7.3 TFLOP at 16 frames and 16.6 at 64, more than our entire escalation (5.2 and 6.1).
At 10\% of the history tokens, the best of three training-free rules is 3.1 points on choice and 4.8 on point behind \HL, and pruning catches up only when it keeps half of the tokens, at twice our escalation compute.
Watching fewer frames is no better: at 35\% more compute the 8B watching 4 frames loses 4.4 points on choice, 16 half-resolution frames lose 4.1 on point, and 8 frames need 76\% more compute to tie \HL.
The writer's caption of the history is a much weaker channel: read zero-shot it scores below the 8B with no history, and after fine-tuning the reader on captions it is still $-$3.7 [$-$4.9, $-$2.5] choice (pre-registered) and $-$5.8 [$-$7.9, $-$3.8] point below \HL, at 41\% more compute.
Text discards visual information that the 8B uses; the KV handoff keeps most of it.

\paragraph{Recomputing the present is enough.}
\label{sec:exp_rarecache}
Our reader recomputes only the current image and the text after it, 36\% of a 16-frame RoboMME prefix.
RaReCache~\cite{sarkar2026rarecache} instead maps everything in closed form and recomputes the positions whose mapped KV carries the most energy outside a calibration subspace, wherever they are.
At the same number of recomputed positions \HL matches it (+0.9 [$-$0.4, 2.3] choice, $-$0.3 [$-$1.9, 1.2] point; pre-registered), and a higher-capacity variant of our map with a nonlinear residual is +2.2 [0.8, 3.6] choice above the RaReCache-style path.
But 39\% of the rank-selected positions lie in the history, so the reader must run its vision tower over it again: at 16 frames the RaReCache-style path costs 6.83 TFLOP against 5.21 for ours, 24\% less at equal accuracy, and the gap widens with the history (\cref{fig:cost_scale}).
Trained through the hybrid path, the map lets recomputation stay at the present.

\setcounter{topnumber}{1}%
\subsection{The memory cascade matches re-watching}
\label{sec:exp_cascade}

\begin{figure}[t]
  \centering
  \includegraphics[width=\linewidth]{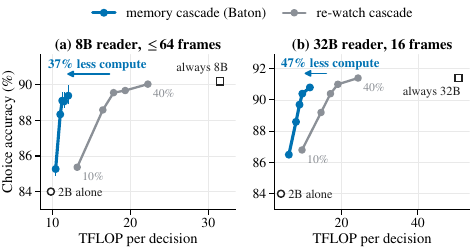}
  \caption{\textbf{Accuracy vs.\ compute of the two escalation channels} (RoboMME choice; budgets of 10--40\%, same router and thresholds). At every budget the memory cascade is within 0.7 points of the re-watch cascade, at 21--46\% less compute for the 8B reader (mean of three seeds) and 36--52\% less for the 32B reader.}
  \label{fig:cascade}
\end{figure}

\begin{table}[t]
  \caption{\textbf{Cascades at a 25\% escalation budget} (accuracy in \%, TFLOP per decision; always-large costs 13.4, 31.5 and 50.6 TFLOP). Memory cascades use \HL after hard-decision training (8B: mean of three seeds).}
  \label{tab:cascade}
  \centering\footnotesize
  \setlength{\tabcolsep}{2.6pt}
  \settowidth{\vaw}{\textbf{11.27}}
  \begin{tabular}{@{}llc>{\columncolor{oursrow}}c>{\columncolor{oursrow}}cccc@{}}
    \toprule
    & & & \multicolumn{2}{>{\columncolor{oursrow}}c}{\textbf{Memory cascade}} & \multicolumn{2}{c}{Re-watch cascade} & Always \\
    Reader & Task & 2B & acc. & TFLOP$\downarrow$ & acc. & TFLOP$\downarrow$ & large \\
    \midrule
    8B, 16 fr. & choice & 84.0 & \textbf{87.8} & \rb{\textbf{5.50}} & 88.5 & \rb{7.59} & 88.8 \\
    & point & 80.5 & \textbf{81.8} & \rb{\textbf{5.44}} & 82.8 & \rb{7.43} & 83.1 \\
    8B, $\le$64 fr. & choice & 84.0 & \textbf{89.1} & \rb{\textbf{11.27}} & 89.6 & \rb{17.86} & 90.2 \\
    & point & 80.4 & \textbf{81.9} & \rb{\textbf{11.10}} & 82.9 & \rb{16.94} & 83.4 \\
    32B, 16 fr. & choice & 84.0 & \textbf{89.7} & \rb{\textbf{9.02}} & 90.4 & \rb{17.11} & 91.4 \\
    & point & 80.5 & \textbf{83.5} & \rb{\textbf{8.81}} & 84.8 & \rb{16.53} & 86.1 \\
    \bottomrule
  \end{tabular}
\end{table}

Escalating through the memory matches escalating by re-watching to within a point, at 28--47\% less compute.
The writer's confidence picks the decisions to escalate (AUROC 0.93 for its errors on choice, 0.90 on point), and a threshold set on validation episodes transfers to test.
Before hard-decision training the memory cascade trails the re-watch cascade by 1.6 [0.6, 2.5] on choice, because escalated decisions are exactly where \HL is weakest; hard-decision training closes this gap (\cref{tab:cascade}, \cref{fig:cascade}).
With the 8B reader the memory cascade is within 0.5--1.0 points of the re-watch cascade at 72\% (16 frames) and 63\% (64 frames) of its compute, stably across three training seeds.
On one RTX 4090, an escalation through the memory takes 0.083\,s against 0.140\,s for re-watching with 16-frame histories, where its peak activation memory falls from 281 to 138\,MiB, and 0.116\,s against 0.305\,s with 64 frames. The map adds 0.56\,GiB to the 16.3\,GiB reader.

\paragraph{The savings grow with the reader.}
With a 32B reader, re-watching costs 50.6 TFLOP per escalation and reading the memory 19.0, so the memory cascade needs 53\% of the re-watch cascade's compute while within 0.7 choice points of it (89.7 against 90.4, $-$0.7 [$-$1.7, 0.3], non-inferior at a pre-registered 2-point margin) and 1.3 points on point.

\subsection{Larger readers, other families, OpenEQA}
\label{sec:exp_scale}

\begin{figure}[t]
  \centering
  \includegraphics[width=\linewidth]{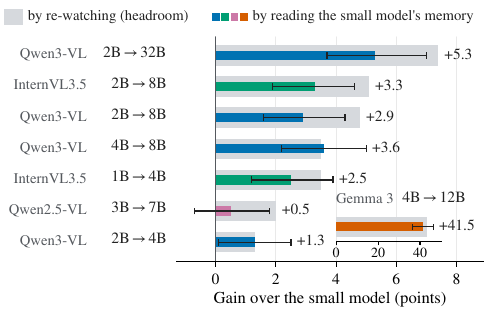}
  \caption{\textbf{The handoff gain tracks the reader's headroom.} Grey: gain of the large model over the small one by re-watching; colour: by reading the small model's memory (whiskers: 95\% intervals, Bonferroni-corrected). On choice the gain follows the headroom closely ($r = 0.91$, seven pairs; inset: Gemma~3, pointing).}
  \label{fig:headroom}
\end{figure}

\begin{table}[t]
  \caption{\textbf{OpenEQA handoff} (LLM-Match; retention in parentheses). Maps are trained on VSI-Bench scenes and transferred without OpenEQA labels; ScanNet was held out until all choices were fixed. Last row: the pre-registered primary comparisons with 95\% intervals (adapted map on HM3D, zero-shot map on ScanNet). \HL is best or statistically tied in every column.}
  \label{tab:openeqa}
  \centering\small
  \setlength{\tabcolsep}{3.5pt}
  \settowidth{\vbw}{\ci{2.8, 10.8}}
  \begin{tabular}{@{}lrlrlrl@{}}
    \toprule
    & \multicolumn{3}{c}{HM3D} & & \multicolumn{1}{c}{\makebox[0pt]{ScanNet}} & \\
    \cmidrule{2-4}\cmidrule{6-6}
    LLM-Match & \multicolumn{1}{c}{\makebox[0pt]{16 fr.}} & & \multicolumn{1}{c}{\makebox[0pt]{64 fr.}} & & \multicolumn{1}{c}{\makebox[0pt]{16 fr.}} & \\
    \midrule
    8B, no history & 37.2 & & 37.2 & & 35.0 & \\
    2B alone & 39.6 & & 46.5 & & 50.4 & \\
    8B re-watches & 48.9 & & 53.9 & & 58.6 & \\
    Text memory & 34.9 & & 37.1 & & 44.4 & \\
    \midrule
    \multicolumn{7}{@{}l}{\grp{8B reads the 2B's memory, no OpenEQA labels}} \\
    \rowcolor{oursrow} zero-shot map & 47.1 & \an{{\scriptsize(0.81)}} & 48.0 & \an{{\scriptsize(0.20)}} & 53.5 & \an{{\scriptsize(0.38)}} \\
    \rowcolor{oursrow} zero-shot \HL & 46.6 & \an{{\scriptsize(0.76)}} & 49.9 & \an{{\scriptsize(0.46)}} & 54.9 & \an{{\scriptsize(0.55)}} \\
    \rowcolor{oursrow} label-free adaptation & 46.4 & \an{{\scriptsize(0.73)}} & 50.2 & \an{{\scriptsize(0.50)}} & 54.2 & \an{{\scriptsize(0.46)}} \\
    \midrule
    Handoff $-$ 2B & \textbf{+6.8} & \an{\ci{2.8, 10.8}} & \textbf{+3.7} & \an{\ci{0.5, 6.9}} & \textbf{+3.1} & \an{\ci{0.4, 5.8}} \\
    \bottomrule
  \end{tabular}
\end{table}

\paragraph{A larger reader reads the same memory better.}
After the same fine-tuning, Qwen3-VL-32B is +7.4 choice and +5.6 point above the 2B.
Reading the \emph{same} 2B memory, it gains +5.3 [3.7, 7.0] choice points over the 2B (pre-registered), recovers 72\% of its headroom, and is +2.4 [1.2, 3.7] above the 8B reading that memory (\cref{fig:headroom}).
It reads the history, not just the present (+7.3 choice and +6.0 point over the 32B with no history), and the distilled map again beats translation, by +4.3 [3.0, 5.5] on choice.

\paragraph{Other model families.}
\label{sec:exp_family}
InternVL3.5~\cite{internvl35} differs from Qwen3-VL in its vision encoder, token layout and positional scheme, and the recipe carries over with only these interface changes.
The 4B reading the 1B's memory gains +2.5 [1.2, 3.9] choice points and the 8B reading the 2B's gains +3.3 [1.9, 4.6] (both pre-registered); on the memory-dependent tasks they are +8.0 and +11.7 above the reader with no history.
Translation fails here too: for 2B$\to$8B it is 3.0 choice points \emph{below} the 2B answering alone and 6.3 [4.8, 7.7] below \HL.
On Qwen2.5-VL 3B$\to$7B~\cite{qwen25vl}, whose 7B barely leads its 3B (+2.0 choice), the handoff has little to recover (+0.5 [$-$0.7, 1.8]) but clearly reads the memory (+6.7 [4.9, 8.7] point over the 7B with no history).

\paragraph{A reader can use what its writer cannot.}
\label{sec:exp_gemma}
Gemma~3~\cite{gemma3} (4B$\to$12B) is the pair least like the others: a language model outside the Qwen line, sliding-window attention in most layers, and half as many key--value heads in the writer as in the reader.
Its fine-tuned 4B knows \emph{what} to do but not \emph{where}: on the decisions that ask for a point it chooses correctly in 91.0\% and points correctly in 32.7\%, against 76.0 for the 12B re-watching.
Reading the 4B's memory, the 12B points at 74.2: +41.5 [36.3, 46.5] over the 4B (pre-registered), +7.5 [5.4, 9.6] over the closed-form map and +7.2 [5.0, 9.4] over the 12B with no history, 80\% of what re-watching adds (+2.1 [0.9, 3.2] on choice).
Of the 879 decisions the 4B points wrongly, the 12B gets 68.0\% right from the 4B's memory and 70.9\% by re-watching.
The cache thus holds what the small model saw, including what it cannot itself act on, and a larger reader recovers it.

\paragraph{Gain follows headroom.}
Across all eight pairs, from four families, the gain follows the reader's headroom over the writer (\cref{fig:headroom}): $r = 0.91$ on choice over the seven Qwen and InternVL pairs. The handoff delivers what the larger model has to offer, and more the more there is.

\paragraph{Zero-shot transfer to OpenEQA.}
\label{sec:exp_openeqa}
The handoff transfers to OpenEQA without OpenEQA training, and label-free adaptation in the target scenes adds nothing (\cref{tab:openeqa}).
On 16-frame HM3D histories it retains 73--81\% of the 8B's advantage and beats the 8B with no history by 9--10 points.
On the held-out ScanNet split it beats the 2B (+3.1 [0.4, 5.8], pre-registered), and on 64-frame histories it retains about half (+3.7 [0.5, 6.9] over the 2B, pre-registered).
A text memory is again much weaker: the zero-shot map beats the 2B's caption by +12.3 [8.4, 16.3] on HM3D and +9.1 [6.1, 12.2] on ScanNet (pre-registered).
On open-ended answers the 2B's confidence barely ranks its errors, so routing, not the memory, is the bottleneck: a learned router passes its pre-registered gate (+2.6 [0.9, 4.4] over the 2B at a 25\% budget) and ties the re-watch cascade with the same router ($-$0.6 [$-$2.0, 0.7]).

\subsection{Why it works: one direction, twelve layers}
\label{sec:exp_mech}

\begin{figure*}[t]
  \centering
  \includegraphics[width=\linewidth]{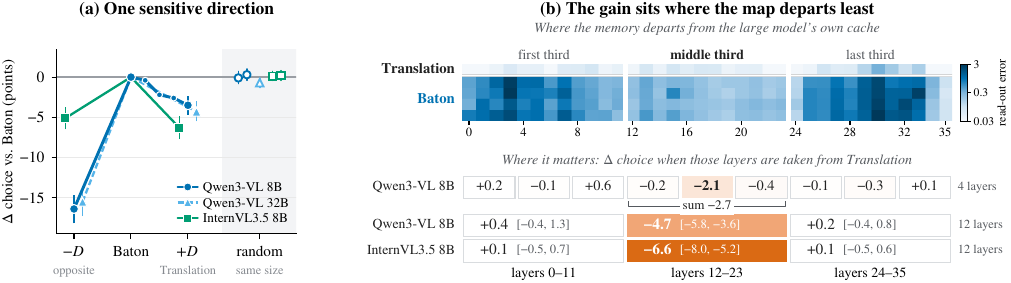}
  \caption{\textbf{What distillation adds is one direction, and it sits where the map departs least} (RoboMME choice). (a) The step $D$ that takes \HL to the translation map costs the gain (small markers: partial steps) and the opposite step far more; random steps of identical size (open markers) cost nothing. (b) Top: read-out error per layer of the 8B: \HL departs from the reader's own cache everywhere, but least in the middle third. Bottom: layers taken from the translation map, four (first row) or twelve at a time, lose the gain only in the middle third; its three windows sum to 2.7, less than the 4.7 of the twelve together: the layers act jointly.}
  \label{fig:mech}
\end{figure*}

\Cref{sec:diag} hypothesized that the reader needs a usable memory, not a faithful one.
Because the map is linear, this can be tested directly.

\paragraph{Fidelity does not predict accuracy.}
Every map distilled through the deployed path reads very differently from the reader's own cache (read-out error 0.48--0.85, against 0.06 for the fitted maps) and is 2.6--5.0 choice points more accurate.
Across 13 trained maps neither the fit error that regression minimizes nor the read-out error predicts accuracy (Spearman $+0.12$ and $+0.09$); only the divergence of the final answer does ($-0.57$).
Heo~\etal~\cite{heo2026crossmodel} find that, for fitted maps, attention-output fidelity, unlike KV fit, predicts transfer across model pairs; across training signals within a pair it does not.
The maps trained with the KV error alone and with KL plus the same KV error differ in fit error by 4\% on keys and 2\% on values, yet the second is 5.0 choice points more accurate: maps that fit the reader's KV almost equally well can carry the whole gain or none of it.
Accordingly, only the answer term of the loss is essential: without it the map is no better than the 2B, with it the choice gain is there whatever the weight of the KV error, and the initialization, the layer pairing and the amount of data matter little (250 decisions give +2.1 [0.8, 3.4]).

\paragraph{A direction, not a distance.}
Mixing two linear maps mixes their outputs exactly, so \HL can be moved in chosen directions by chosen amounts (\cref{fig:mech}a).
Let $D$ be the translation map, fitted to the KV error, minus \HL.
Moving \HL along $D$ to the fitted map costs 3.5 [2.3, 4.7] choice points.
Replacing $D$ by $DQ$, with $Q$ a random orthogonal matrix on the output dimension, perturbs every mapped token by exactly the same norm in a random direction: it costs nothing ($-$0.1 [$-$0.8, 0.7] and +0.3 [$-$0.4, 1.0] for two draws), although it makes the map \emph{less} faithful than \HL by every measure.
The same step along $-D$ is catastrophic: $-$16.4 [$-$18.2, $-$14.6].
The reader is thus indifferent to a generic perturbation of this size and highly sensitive to this one axis: KV regression leaves the map close in norm but displaced along the one direction that matters, with no signal that would move it.
The three steps separate in the same way for a 32B reader ($-$4.3, $-$0.7 and $-$15.5) and for InternVL3.5 ($-$6.3, +0.1 and $-$5.1), where only the collapse beyond \HL is absent.

\paragraph{Twelve layers carry the whole gain.}
Layers have separate weights and can be exchanged between the two maps (\cref{fig:mech}b).
With the fitted map in the first or the last third of the reader's 36 layers and \HL elsewhere, nothing is lost on choice (+0.4 and +0.2); with the fitted map in the middle third alone the handoff loses 4.7 [3.6, 5.8] points.
Conversely, the fitted map with \emph{only} layers 12--23 taken from \HL is as accurate as \HL itself (+0.5 [$-$0.4, 1.3]).
The middle third is thus both necessary and sufficient, and the same holds for InternVL3.5 ($-$6.6 and +0.3 [$-$0.3, 1.0]).

\emph{The twelve act jointly.}
Replacing four consecutive layers at a time (\cref{fig:mech}b, first row of the strip) changes choice by at most 0.6 points outside layers 12--23; inside, layers 16--19 cost 2.1 [1.2, 3.0] and their two neighbours 0.2 and 0.4.
The three sum to 2.7 points, well short of the 4.7 lost when the twelve are replaced together (difference 1.9 [0.6, 3.4]): no four layers hold the gain on their own.

\emph{The gain sits where \HL moved least.}
Layer by layer, its read-out departs from the reader's own at every depth, but in the middle third the error is 0.23--0.27 in each of the four runs, against 0.67--0.89 in the first third and 0.80--1.23 in the last.
Window by window the picture is the same (\cref{fig:mech}b, top against bottom): layers 16--19, the most costly to replace, are among the least changed (0.21--0.22), and layers 28--31, the most changed (1.2--2.3), can be replaced for free.
The large departures in the outer layers are the ones the reader tolerates, and the small one in the middle carries the gain: across depth, as across directions, the size of a change does not say whether it matters to the reader.

\emph{The correction is mostly in the keys}, in where the reader looks: fitted keys in the middle layers cost the whole gain ($-$3.2 [$-$4.1, $-$2.2], as much as fitted keys in every layer) and fitted values 1.8 [1.0, 2.6], and the distilled keys of these twelve layers alone lift the fitted map half of the way to \HL.
On its own cache the reader places 0.70 of its attention mass on the history and on the fitted maps almost the same (0.66--0.67); on \HL it places 0.56 (less in all 200 decisions), leaning more on the frame it encodes and taking from the history selectively. Distillation thus changes where the reader looks in the history, and how much, more than what it retrieves there.

\newpage
\paragraph{Consequences.}
The reader can use memories that are not its own, but only some of them, and a regression target does not say which.
This is why the map must be trained on the reader's answers through the deployed path, why a linear map suffices (what the closed-form solution lacks is not capacity but a direction), and why the choice gain is stable across training runs although the maps themselves differ: there are many usable memories.
\section{Conclusion}
\label{sec:limitations}

A small VLM's KV cache over a video history is a memory that a larger VLM of the same family can read.
With \HL, a per-layer linear map distilled from the large model's own answers through the deployed path, the large model reads that memory better than the small model can use it, and better than a translated cache, a pruned or shortened re-watch at higher cost, or a text memory.
As the escalation channel of a cascade it matches re-watching to within about a point at 28--47\% less compute.
What the map must deliver is a usable memory, not a faithful copy: fitting the reader's cache more closely does not help, while distillation finds the one direction, in the middle third of the reader's layers, that does.
Escalation no longer has to mean perceiving the history again.

\paragraph{Limitations.}
Localization transfers less readily than semantics: point gains are smaller than choice gains for every pair but Gemma~3, and closing this gap is left to future maps.
Each writer--reader pair has its own map, trained within one model family, and we study histories of up to 64 frames; longer streams, where re-watching costs most, and one map shared across readers are natural next steps.
{
    \small
    \bibliographystyle{ieeenat_fullname}
    \bibliography{main}
}

\end{document}